# An Effective and Efficient Method for Detecting Hands in Egocentric Videos for Rehabilitation Applications

Ryan J. Visée, Jirapat Likitlersuang, *Graduate Student Member, IEEE* and José Zariffa, *Senior Member, IEEE*

*Abstract*— *Objective:* **Individuals with spinal cord injury (SCI) report upper limb function as their top recovery priority. To accurately represent the true impact of new interventions on patient function and independence, evaluation should occur in a natural setting. Wearable cameras can be used to monitor hand function at home, using computer vision to automatically analyze the resulting videos (egocentric video). A key step in this process, hand detection, is difficult to accomplish robustly and reliably, hindering the deployment of a complete monitoring system in the home and community. We propose an accurate and efficient hand detection method that uses a simple combination of existing detection and tracking algorithms.** *Methods:* **Detection, tracking, and combination algorithms were evaluated on a new hand detection dataset, consisting of 167,622 frames of egocentric videos collected from 17 individuals with SCI performing activities of daily living in a home simulation laboratory.** *Results:* **The F1-scores for the best detector and tracker alone (SSD and Median Flow) were 0.90±0.07 and 0.42±0.18, respectively. The best combination method, in which a detector was used to initialize and reset a tracker, resulted in an F1-score of 0.87±0.07 while being two times faster than the fastest detector alone.** *Conclusion:* **The combination of the fastest detector and best tracker improved the accuracy over online trackers while improving the speed over detectors.** *Significance:* **The method proposed here, in combination with wearable cameras, will help clinicians directly measure hand function in a patient's daily life at home, enabling independence after SCI.**

*Index Terms*— **Computer vision, Egocentric, Object detection, Spinal Cord Injury, Upper limb rehabilitation**

## I. INTRODUCTION

CERVICAL spinal cord injuries (SCI) significantly reduce the quality of life of the affected individuals and entails an estimated economic cost of $2.7 billion per year in Canada [1]. In particular, the impairment of arm and hand function

This work was supported in part by the Natural Sciences and Engineering Research Council of Canada (RGPIN-2014-05498), the Rick Hansen Institute (G2015-30), and the Ontario Early Researcher Award (ER16-12-013).

R. Visée and J. Zariffa are with KITE – Toronto Rehab – University Health Network, Toronto, ON, CA and the Institute of Biomaterials and Biomedical Engineering (IBBME), University of Toronto, Toronto, ON, CA. J. Zariffa is also with the Edward S. Rogers Sr. Department of Electrical and Computer Engineering, University of Toronto, Toronto, ON, CA and the Rehabilitation Sciences Institute, University of Toronto, Toronto, ON, CA. J. Likitlersuang is now with Harvard Medical School, Harvard University, Cambridge, MA, USA. (correspondence e-mail: jose.zariffa@utoronto.ca).

plays a major role in the loss of independence after SCI. Individuals with cervical SCI report upper limb function as their top recovery priority [2]. As a result, new treatments to improve hand function after SCI are needed. Current assessments of the severity of upper limb impairments are typically performed in clinical settings. To accurately represent the true impact that new interventions have on patient function and independence, evaluation should occur at home. Currently, there are no methods that directly measure and track the effect of therapy on patient hand function in their daily life at home.

With the emergence of wearable cameras, such as Google Glass™ and GoPro®, innovative ways to directly measure hand function at home in persons with SCI have become available. In fact, wearable cameras are already being used to collect data and evaluate human interactions [3]-[6]. Wearable cameras are of interest as they capture activities from the camera-wearer´s point of view, which can be used to understand daily activities such as meal preparation and other functional self-care tasks. First-person cameras also allow for large data collection with fewer limitations compared to fixed cameras which are limited to one location, resulting in data loss and occlusions, along with inaccurate representations of daily activities. Home rehabilitation is of utmost interest as the natural movement information provided by wearable cameras can be used to monitor patient performance and independence in activities of daily living (ADLs), and provide feedback for more effective and more accessible rehabilitation.

Although videos from wearable cameras (egocentric videos) can be used to monitor patient activities at home, the automated analysis of egocentric videos using computer vision presents significant technical challenges [5]-[6]. A problem exists in the detection of hands in egocentric videos, which is a necessary first step prior to hand function analysis. Robustly and reliably detecting and tracking the hand is affected by factors including partial occlusions, lighting variations, hand articulations, camera motion, and background or objects that are similar in color to the skin.

In addition, computationally efficient solutions to this problem are desirable as a step towards a system capable of real-time video processing, which would reduce privacy concerns by avoiding the need to store raw videos for later analysis.

Therefore, this study aimed to generate an algorithm for fast and reliable hand detection in egocentric videos captured by individuals with cervical SCI by finding the best trade-off between accuracy and speed. We integrated object detection techniques with tracking algorithms, proposing a method that can increase the computational efficiency of hand detection algorithms with competitive accuracy in egocentric videos compared to previous approaches.

## II. RELATED WORK

### A. Wearable sensors for healthcare purposes

Clinical assessments such as the Graded Redefined Assessment of Strength Sensibility and Prehension (GRASSP) and Spinal Cord Independence Measure (SCIM) normally occur within clinical settings or rely on self-report, and do not directly capture the true impact of interventions on a person in their daily life at home [7]-[8]. It is therefore important to develop tools that can measure an individual's function directly at home. As a result, research on wearable sensors for rehabilitation applications has increased in popularity. Previously used physical sensor systems include goniometers, accelerometers, piezoelectric pressure sensors, flexible sensors, and inertial sensors [9]-[11]. The most common approach for monitoring upper limb function has been to use wrist-worn accelerometers [12]-[14]. However, this approach is better suited to detecting arm movements and may not capture finer movements associated with dexterous hand use. Due to the large number of degrees of freedom, the potential for variations in sensor placement, and number of different hand behaviors, wearable sensor systems for the hand are far less developed compared to sensors used on other areas of the body [11]. Specifically for hand function, mechanical glove systems, magnetic rings, and finger-worn accelerometers have been proposed [15]-[17], but further study will be required to establish the viability of these systems in unconstrained environments and tasks. Egocentric video is appealing in this context because it can capture information not only about the hand itself but also about its interactions with the environment [6], [18].

### B. Object or Hand Detection

To analyze hand function in egocentric videos, it is important to first detect hands. Recent work by Betancourt et al. [19] and Bambach et al. [4] showed the importance of a hand detection step before further analysis such as hand segmentation. Hand detection is a specific application of a more general and fundamental problem in computer vision, known as object detection. Recently, significant progress has been made in improving the performance of object detection using convolutional neural networks (CNNs). Existing algorithms can be divided into two categories, region-based and regression-based approaches. Region-based approaches generate a set of region or object proposals in an image and then perform classification on each proposal. This approach was applied notably in the region-based CNN (R-CNN) but suffered from expensive computational costs as the region proposals must be calculated and classified in every frame [20]. To improve the speed, Faster R-CNN was introduced, which increased both speed and accuracy but still performed well-below real-time (defined here as 30 frames per second (FPS)) [21]. This algorithm was applied specifically in hand detection but generated region proposals in areas of an image in which the hand would most likely appear, increasing both the efficiency and accuracy of hand proposal generation [4]. Regression-based approaches implement algorithms that can directly predict the location of bounding boxes rather than classify object proposals. You Only Look Once (YOLO) is one algorithm that uses a single CNN to simultaneously predict bounding boxes and class probabilities, competitively performing with Faster R-CNN, while being significantly faster [22]. Subsequently, the second version of YOLO (YOLOv2) outperformed Faster R-CNN in both accuracy and speed while performing in real-time [23]. Another regression-based algorithm that outperformed Faster R-CNN was the Single-Shot Multibox (SSD) Detector [24]. The SSD framework is similar to YOLOv2 in design but consists of visualizing an image using feature maps at different aspect ratios in convolutional fashion.

### C. Object Tracking

Object detection techniques are limited by the long computational costs and the inability to associate detections over frames. In contrast, tracking algorithms aim to save the identity of the object and predict the new location of the object in the next frame based on dynamics and previous frame information. This allows tracking algorithms to perform faster than detection algorithms, making them a desirable tool for real-time applications. However, tracking algorithms have difficulty recovering from occlusions and can accumulate errors over time, resulting in the tracker drifting away from the object and reducing applicability in object detection tasks.

Online learning algorithms are not pre-trained on any specific dataset but are instead given a single image and a manually selected bounding box as an initial ground-truth. They attempt to learn a model based on an object's appearance with past and present examples extracted from a video [25]-[26]. One of the more powerful trackers, the Kernelized Correlation Filter (KCF) tracker, exploits the power of Fourier analysis and circulant matrices by working in the dual space using the kernel trick [27]. Finally, the Median Flow (MF) tracker tracks the object both forward and backward in time using Forward-Backward error, a simple measure of the difference between the forward and backward trajectories [28]. These systems could be made more robust by training the trackers offline on a large dataset [29]. However, the top offline algorithms are not feasible to deploy on portable devices due to their slow processing time and report similar accuracy to the KCF tracker [29].

### D. Combining Object Detectors and Trackers

For more complex situations such as multi-person tracking, detecting and tracking individuals is more complicated as individuals can be occluded for long periods. Also, with many people in a single scene, it is difficult to associate person detections between frames to a specific individual. Therefore, the ability to associate certain detections with certain tracked targets is applied in approaches known as tracking-by-detection. However, these methods use the detector and tracker simultaneously, increasing the complexity of the system and reducing performance time [30]-[32]. Most comparable to our work is Bu et al. [33], who used a combination of the Faster R-CNN detector and the KCF tracker for multi-object tracking in third-person videos. This approach also performed detection and tracking in every frame and then compared the state of each to obtain the correct location of the object. While these simultaneous computations may be needed for multi-object detection, we show that a system that focuses on one type of object does not require such complexity.

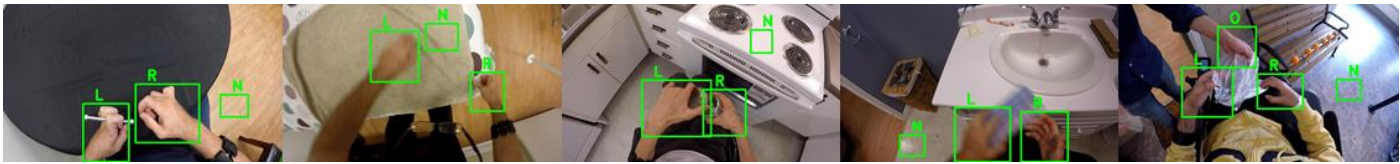

Fig. 1. Example annotated frames in the ANS SCI hand detection dataset.

## III. METHODS

### A. Egocentric Hand Detection Dataset

The egocentric hand detection dataset used for this study was obtained from previous experiments that resulted in videos collected using wearable cameras on individuals with SCI, termed the Adaptive Neurorehabilitation Systems (ANS) SCI dataset [6]. The ANS SCI dataset contains 17 individuals with SCI performing a variety of ADLs, collected in a home simulation laboratory at the Toronto Rehabilitation Institute – University Health Network. Videos in this dataset were recorded using a head-mounted GoPro® HERO 4 wearable camera recorded at 30 FPS with 1080p resolution. This dataset represents ADLs in many different environments, including the kitchen, washroom, living room, dining room, bedroom, and hallway. Participants were asked to manipulate over 30 objects in over 35 ADLs as naturally as possible. Participants were not specifically asked to hold hands in view of the camera and were not given specific instructions on how to perform ADLs. Therefore, the ANS SCI dataset reflects a range of objects, environments, ADLs, and participants, including different levels of impairment.

We generated a large hand detection dataset (Fig. 1) by manually labeling bounding boxes around hands in a subset of frames covering every participant, ADL, and environment. The complete dataset consists of 167,622 images containing labels for "left hand"/"right hand" (L/R), which belong to the camera-wearer, and "other hands" (O), which belong to anyone else that may appear within the video. It also contains labels for "not hand" (N), which was used as negative data to generate labels for objects and background in areas that the CNN may confuse as hands. Images and bounding box annotations are at a resolution of 720x405. Care was taken to ensure a large distribution between participants, ADLs, and environments, while also including many difficult annotations such as occlusions, impaired hand postures, and quick movements.

### B. Detection and Tracking Only

This work built upon previous detection and tracking algorithms that were made to fit the hand detection problem.

For hand detection, we implemented Faster R-CNN [21], YOLOv2 [23], and SSD [24]. These models were trained using the ANS SCI dataset with minor modifications to hyperparameters. Although Bambach et al. [4], who used a region-based approach and introduced a more efficient hand-proposal generation method, showed great potential in hand detection for egocentric videos across different participants and environments, this algorithm was not specifically implemented using our ANS SCI dataset. However, we do compare our proposed algorithm to theirs in Section IV.C.

For hand tracking, we implemented 4 online tracking algorithms due to their efficiency on CPU processors; Online Boosting (OLB), Multiple Instance Learning (MIL), KCF, and MF [34], [26]-[28]. Although online trackers are not robust to challenging situations, such as occlusions or fast motions, offline trackers, which would benefit from our large dataset, are not feasible to deploy on portable devices due to their slow processing time. Also, the KCF tracker has reported similar accuracy to these offline approaches, while being significantly faster on a CPU [29]. Therefore, we did not implement offline trackers despite their high accuracy, as the proposed combined algorithm would not benefit in efficiency.

### C. Combining Object Detectors and Trackers

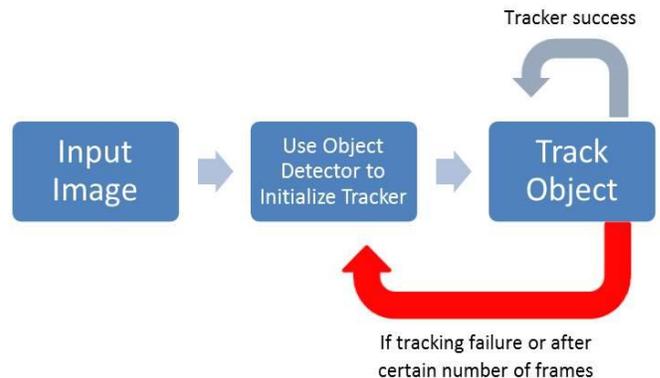

Fig. 2. Proposed Detector-Assisted Tracking (DAT) pipeline

Similar to tracking-by-detection algorithms, we proposed the use of an object detector to automatically initialize and reinitialize an object tracker upon failure or after a certain number of frames (Fig. 2). This method was proposed since the main problem with tracking algorithms is the inability to recover from occlusions or lost objects, thus making it difficult to perform adequately after failure. Therefore, we aid successful recovery from occlusions and quick motions by using a detector. Further, since online trackers require manual initialization, the process is only semi-automatic. Using a detector to initialize the tracker fully automates the process. Another problem many online trackers face is tracker drift. Using a detector to reset the tracker after a certain number of frames minimizes the effect of tracker drift, thus avoiding the propagation of errors and improving performance. We refer to this proposed method as "Detector-Assisted Tracking" (DAT).

This proposed method is most similar to Bu et al. [33], who used a combination of the Faster R-CNN detector and the KCF tracker for multi-object tracking in third-person videos. However, they performed detection and tracking simultaneously in every frame and then compared the state of each to obtain the correct location of the object. In contrast, we only use the detector to initialize the tracker at the beginning of a video and to reinitialize the tracker when it fails or after a certain number of frames. Therefore, either the detector or tracker is used to determine the hand location in a certain frame but not both. This minimized the required detections, thus improving the accuracy over trackers-alone while maintaining the efficiency of these approaches. To further minimize detector usage, the tracker was disabled if it failed and the detector was unable to locate the hand in a certain number of consecutive frames. The detector then checked once every certain number of frames until the hand was found. The performance was based on the accuracy and processing time of the tested trackers with and without the aid of a detector.

Parameters tested were defined as reset iterations, consecutive intersection over unions (IOU), and check iterations. Reset iterations is the number of frames between each detector usage to reinitialize the tracker and combat against tracker drift. If this parameter was 100, then the detector would be used every 100 frames to reinitialize the tracker or any time the tracker failed. Consecutive IOU is the number of consistent detections used to initialize the tracker. If consecutive IOU was 3, then the tracker would be initialized only if the detector found the hand in 3 consecutive frames and every detection had an overlap greater than 0.1 with the previous detection. This assumes that false positives would not be detected consistently across frames. This step also assumes that hands will not move a considerable amount over consecutive frames, hence the 0.1 overlap threshold. The consecutive IOU parameter was also used to disable the tracker if it did not successfully find the hand in the set number of consecutive frames. Finally, check iterations is the number of frames after the tracker was disabled in which the detector attempted to locate the hand. If check iterations was 60, then every 60 frames after the tracker was disabled the detector checked to see if the hand existed. If in that 60th frame the detector was able to locate the hand then the detector attempted to reinitialize the tracker. The tracker remained disabled if the detector was unable to locate the hand. Disabling the tracker was used to improve efficiency by ensuring neither the tracker nor detector was being used during periods in which the hand was not in the video. Combinations are referred to as "resetIterations/consecutiveIOU/checkIterations" and would be 100/3/60 for the example provided above.

This work builds upon a feasibility study performed by Visée et al. [35] which reported that on a subset of the ANS SCI hand detection dataset, the best combination resulted in a 1.7x improvement in F1-score compared to the best tracker alone (MF) and was 3x faster than the fastest detector alone (YOLOv2) on a CPU. This resulted in the conclusion that DAT would be a feasible combination method.

## D. Evaluation Method

To account for participants' functional capabilities, ADLs, environments, and variability, the dataset was split into 3 groups to generate balanced training and testing sets for a cross-validation process. The split was based on participants and we used the International Standards for Neurological Classification of Spinal Cord Injury (ISNCSCI) assessment tool to account for hand function [36] (Table 1). We specifically used the upper extremity motor subscore (UEMS) to divide our dataset since our focus is on hand function. To generate UEMS scores, 5 upper limb muscles were manually tested, one from each respective segment of the cervical cord and were scored on a 5-point strength grading scale. The final scores were summed to obtain the total UEMS score. We cycled through these groups by training on 2 subsets and testing on the other, resulting in 3 different trained models. The muscle strength was an important consideration for the dataset split as it ensured one group did not contain more participants with low functional capability or impaired hand posture than the others. This could have resulted in skewed poor performance. Since we considered the participants' muscle strength, we were able to generate a more evenly distributed dataset split with minimal bias. Using a one-way analysis of variance (ANOVA), the means in Table 1 were found to not be statistically different, $F(2,14) = 0.12$, $p = 0.89$.

TABLE 1
ANS SCI DATASET SPLIT BASED ON PARTICIPANTS UEMS.

|  | GROUP A | GROUP B | GROUP C |
|---|---|---|---|
| Average UEMS | $17.83 \pm 5.04$ | $18.80 \pm 3.96$ | $19.00 \pm 4.10$ |
| Total Frames | 63102 | 36051 | 68469 |

Following analysis on our ANS SCI dataset, we tested the generalizability of DAT on two public egocentric hand detection datasets, EDSH [37] and EgoHands [4].

The final performance of hand detection was evaluated using the F1-score on the test set, which is the harmonic mean of precision and recall. The determination of a correct prediction was based on the IOU, which is a measure of the overlap between the predicted bounding box and the ground truth bounding box. In these experiments, we chose an IOU of 0.5 to be an accurate prediction, based on the PASCAL Visual Objects Classes (VOC) challenge [38]. We also considered an IOU between 0.15 and 0.5 to be a correct prediction but with localization error, determined empirically. An IOU score below 0.15 was classified as a background error. In images where more than one detection existed per class, we only considered the bounding box with the highest confidence, as we assume that only one hand type (left or right) can exist for the camera-wearer.

The frame rate of the model was also used as an evaluation metric as the system will ideally run in real-time. For rehabilitation application purposes, a target of 15-20 FPS would most likely provide the same information as a system that runs at the definition of real-time (30 FPS). For real-time information provided in the home and community, these FPS targets should be achieved on mobile CPU processors.

## IV. RESULTS

For final evaluation on ANS SCI, the F1-scores for "left hand" and "right hand" (averaged over all participants within the model's test set) were averaged over the 3 folds of cross-validation to achieve the final scores (Tables 2-3, Fig. 3). The entire ANS SCI dataset and all GPU results were evaluated on a NVIDIA Titan Xp™ 12 GB RAM GPU (Tables 1-5).

### A. Detection and Tracking Only

The results for the 3 implemented object detectors are displayed in Table 2. Detector CPU performance was evaluated on an Intel Core™ i7-8700k™ CPU (CPU-i7). Faster RCNN and SSD were run in Caffe while YOLOv2 was built and run entirely in C/C++, all from the original source.

TABLE II
RESULTS OF DETECTION ALGORITHMS ON ANS SCI DATASET

| Algorithm | F1-SCORE | FPS ON GPU | FPS ON CPU-i7 |
|---|---|---|---|
| SSD | 0.90 ± 0.07 | 44 | 0.5 |
| Faster RCNN | 0.89 ± 0.06 | 15 | 0.4 |
| YOLOv2 | 0.88 ± 0.07 | 68 | 1.5 |

The online trackers (implemented via OpenCV in Python) were tested on the entire ANS SCI dataset (Table 3). Trackers were manually initialized in the first "good" frame in which the hand was seen, chosen empirically, for each video sequence. The CPU FPS rates only were obtained from a subset of the ANS SCI dataset consisting of 19,683 frames but are indicative of the speed on the entire dataset, and evaluated on an Intel Core™ i5-7200U™ CPU (CPU-i5). The F1-scores and GPU FPS rates were evaluated on the entire dataset. Due to the efficiency of online trackers, evaluation was not performed on a GPU.

TABLE III
RESULTS OF ONLINE TRACKERS ON ANS SCI DATASET

| Algorithm | F1-SCORE | MAP | RECALL | FPS ON CPU-i5 |
|---|---|---|---|---|
| MF | 0.42 ± 0.18 | 0.42 ± 0.20 | 0.44 ± 0.19 | 155 |
| KCF | 0.32 ± 0.18 | 0.70 ± 0.27 | 0.24 ± 0.16 | 70 |
| MIL | 0.35 ± 0.14 | 0.31 ± 0.14 | 0.40 ± 0.15 | 17 |
| OLB | 0.31 ± 0.13 | 0.27 ± 0.13 | 0.36 ± 0.14 | 25 |

### B. DAT on ANS SCI

YOLOv2 [23] was the assisting detector used due to its high accuracy performance and efficiency on a GPU (Table 2). As discussed, parameters tested were defined as reset iterations, consecutive IOU, and check iterations. These parameters were initially tested on the subset used by Visée et al. (19,683 frames spanning 6 participants and 4 environments) [23]. We found that although an increase in the reset iterations resulted in slightly faster combinations, it came at a large cost to the F1-score. We found the opposite for consecutive IOU, as increasing this parameter resulted in more accurate combinations with a slight cost in speed. Finally, increasing check iterations resulted in less accurate combinations with no noticeable effect on the speed. Based on the results obtained from the subset of the ANS SCI dataset, we picked 3 models that resulted in the best trade-offs in F1-scores and FPS rates and evaluated them on the full ANS SCI dataset. Note that the DAT method was evaluated on one hand at a time like online trackers, and that the speeds are the average between the two classes, averaged over the 3 folds of cross-validation. The top combinations (Fig. 3) were: 100/3/60, 100/9/60, and 200/8/30. The most accurate model, when averaged over the 3 folds, was YOLO_KCF – 200/8/30 with an F1-Score of 0.87 ± 0.07 and an FPS rate of 133 FPS. The fastest model was YOLO_MF – 100/3/60 with an FPS rate of 283 FPS and an F1-score of 0.81 ± 0.09. Table 4 compares the processing times between YOLOv2 alone and the combinations, for the fastest models. DAT was implemented using a Python wrapper for YOLOv2 and OpenCV in Python for the online trackers.

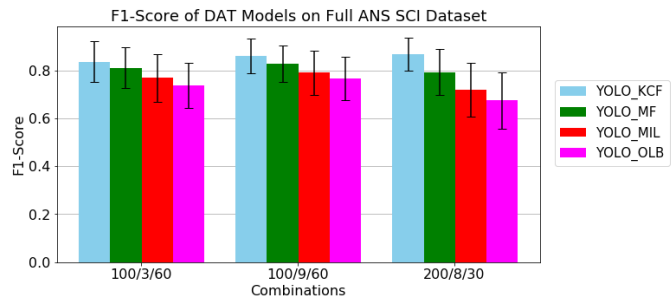

Fig. 3. DAT F1-Score for different combination models and trackers on the entire ANS SCI dataset.

TABLE IV
FPS RATES OF YOLOV2 AND DAT ON ANS SCI DATASET

| Model | YOLOv2 | YOLO_MF | YOLO_KCF | YOLO_MIL | YOLO_OLB |
|---|---|---|---|---|---|
| GPU | 68 | 283 | 166 | 53 | 56 |
| CPU-i5 | 0.3 | 5.5 | 4.4 | 4.5 | 4.5 |

### C. DAT on Publicly Available Datasets

We tested our DAT method on two publicly available detection datasets, EDSH [37] and EgoHands [4]. We report (Table 5) results on each dataset for YOLOv2 combined with the MF and KCF tracker for 100/9/60, as it provided the best trade-off between F1-score and FPS. The images in EDSH and EgoHands were not resized during evaluation and were therefore analyzed at 640x360 and 1280x720 respectively. EDSH was evaluated on 733 frames (converted from pixel-level segmentations to bounding boxes) and EgoHands was evaluated on 800 frames as described in Bambach et al. [4].

TABLE V
DAT PERFORMANCE ON PUBLICLY AVAILABLE DATASETS FOR 100/9/60

| Dataset | YOLO_KCF | FPS | YOLO_MF | FPS |
|---|---|---|---|---|
| EDSH | 0.90 ± 0.05 | 115 | 0.83 ± 0.08 | 205 |
| EgoHands | 0.58 ± 0.28 | 34 | 0.54 ± 0.27 | 65 |

## V. DISCUSSION

High-quality, meaningful outcome assessments are essential to support the development of new treatments to improve hand function after cervical SCI. Despite this need, there are no available methods that directly measure and track the impact of therapy on patient hand function in their daily life at home. Egocentric video is a promising avenue to fill this gap but fully automated analysis is technically challenging. To automatically quantify the functional use of the hand in egocentric videos, we must first determine the correct location of the hand in each frame. Further, to support the use of these techniques in the community, evaluation needs to be computationally inexpensive and deployable on a portable system. In this study, we introduced an effective and efficient algorithm for hand detection by combining existing object detectors and trackers. The competitive accuracy would provide similar information as object detectors alone while the increased speed would result in a system deployable in non-clinical settings for real-time analysis in rehabilitation applications.

We found that all detection algorithms performed with similar F1-score and that the main difference existed in the speed of the systems. YOLOv2 performed the fastest on a GPU at 68 FPS while Faster R-CNN performed the slowest at 15 FPS. However, due to slow speeds on CPU-i7 (less than 1.5 FPS), detectors alone were found to be insufficient for portable systems. On the other hand, all online tracking algorithms were not robust to occlusions or quick motions and therefore suffered in the hand tracking paradigm. We also found that online trackers were highly dependent on user initialization and video quality, resulting in large standard deviations in F1-score. MF obtained the highest F1-score at $0.42 \pm 0.18$ and was also the fastest tracker at 155 FPS. Therefore, online tracking algorithms alone were also insufficient for hand detection in egocentric videos due to their inability to recover from occlusions and quick motions. We showed that combining relatively fast detectors with relatively accurate trackers minimized the faults of each approach resulting in accurate and efficient hand detections.

Based on the results obtained from detectors and trackers alone, we expected a combination between YOLOv2 and MF or KCF to perform the best. Even though KCF performed poorly on its own, it had the potential to perform well upon reset due to its high precision (Table 3). After evaluation, we found this to be true as YOLO_KCF became the most accurate combination, outperforming YOLO_MF. The most accurate combination (YOLO_KCF – 200/8/30) performed 2x better than the best tracker alone (MF) while being 2x faster than the fastest detector alone (YOLOv2) on a GPU (133 vs. 68 FPS). The fastest combination (YOLO_MF – 100/3/60) was 4x faster than YOLOv2 (283 vs. 68 FPS) while still being twice more accurate than MF alone. Therefore, combining detection and tracking algorithms resulted in successful recovery from occlusions and quick motions while improving the speed over detectors alone.

The combinations of YOLO with KCF, MIL, and OLB all performed with similar FPS rates on CPU-i5. This is because MIL and OLB do not report tracking failures and therefore required fewer detections compared to the KCF and MF combinations, increasing the speed of these combinations at the cost of accurate tracks. However, the KCF and MF trackers alone are much faster than MIL and OLB (Table 3), which is why their combinations can still perform fast even though they require more detections. Also, the MF and KCF trackers get a larger boost on a GPU compared to MIL and OLB, resulting in the much greater speed performance on a GPU compared to CPU-i5 (Table 4). To add to the benefits, the combinations displayed lower standard deviation compared to trackers alone, showing that the addition of a detector makes the system more robust and reliable.

The speed increase in these systems, while being almost as accurate as detectors alone, can prove to be beneficial for deployment into public settings. For example, on a less powerful CPU-i5, YOLOv2 ran at 0.3 FPS while YOLO_MF and YOLO_KCF ran approximately 18 and 15 times faster, respectively (5.5 and 4.4 FPS). Although on CPU-i5 we were unable to reach our target of 15-20 FPS, on the more powerful CPU-i7, where YOLOv2 runs at 1.5 FPS, we estimate that YOLO_MF and YOLO_KCF could perform at 20 FPS and 14 FPS respectively, which would meet our goal. This was a limitation of our study as we were unable to force these trackers to only use the CPU on CPU-i7. However, even on a mid-range CPU-i5, we see a significant increase in speed compared to detectors alone.

Testing DAT on two publicly available datasets, EDSH and EgoHands, we first see that DAT generalizes well to EDSH. This shows DAT's ability to generalize to outdoor data even though our dataset contained no outdoor examples. Secondly, upon first glance, it may look as if DAT performs poorly on EgoHands, but our average precision on this dataset is 0.722, which is better than Bambach et al.'s 0.684 when considering only the camera-wearer's hands [4]. This is promising since EgoHands focuses on social interactions rather than on hand detection and therefore contains "other hands" in most frames, which we did not include in our evaluation due to lack of "other hands" examples in our ANS SCI hand detection dataset. In fact, the EgoHands dataset contains the partner's hands in 94.6% of the frames compared to only 62.2% for the camera-wearer's hands. This is in contrast to ANS SCI where "other hands" are only in 4.4% of the frames compared to 71.5% for the camera-wearer's hands. Also, the camera-wearer's hands in EgoHands are only in the videos for short sequences impeding the tracker's ability to learn as it is not given many positive examples.

All detection-by-tracking algorithms mentioned in Section II.D used the detector and tracker simultaneously, increasing the complexity of the system and reducing performance time. While this may be needed for multi-object detection, generating a system that focuses on one type of object does not require a complex approach. Therefore, our novel yet simple approach of either using the detector or tracker, but not both at the same time, resulted in an easy, accurate, and fast algorithm.

## VI. CONCLUSIONS

We have presented a system for effective and efficient hand detection in first-person video. We evaluated this system on the largest known egocentric hand detection dataset, totaling 167,622 frames. DAT, which allows for robust and reliable hand detection while being efficient on a CPU, will aid in the process of evaluating the true impact of new treatments on the lives of persons with SCI, as well as other rehabilitation applications involving hand function. On a CPU, DAT's most accurate method is 2x more accurate than the best tracker alone (MF) while being 15x faster than the fastest detector alone (YOLOv2). Hand detection is an essential step before further analysis can be conducted, including hand segmentation, activity recognition, interaction detection, or grip posture analysis. The development of an ideal hand detection method in combination with the availability of wearable cameras will put researchers one step closer to innovating ways to directly measure hand function in a patient's daily life, thus helping restore independence after SCI.